\documentclass[10pt,twocolumn,letterpaper]{article}

\usepackage[pagenumbers]{iccv} 

\usepackage{graphicx} 
\usepackage{amsmath}
\usepackage{multirow}
\usepackage{algorithm, algpseudocode}

\definecolor{iccvblue}{rgb}{0.21,0.49,0.74}
\usepackage[pagebackref,breaklinks,colorlinks,allcolors=iccvblue]{hyperref}
\usepackage[accsupp]{axessibility}

\title{TAIGen: Training-Free Adversarial Image Generation via Diffusion Models}

\author{
Susim Roy$^{1,2}$, Anubhooti Jain$^2$, Mayank Vatsa$^2$, Richa Singh$^2$ \\
$^1$University at Buffalo, $^2$IIT Jodhpur \\
{\tt\small susimmuk@buffalo.edu}, {\tt\small \{jain.44, mvatsa, richa\}@iitj.ac.in}
}

\begin{document}
\maketitle
\begin{abstract}
Adversarial attacks from generative models often produce low-quality images and require substantial computational resources. Diffusion models, though capable of high-quality generation, typically need hundreds of sampling steps for adversarial generation. This paper introduces TAIGen, a training-free black-box method for efficient adversarial image generation. TAIGen produces adversarial examples using only 3-20 sampling steps from unconditional diffusion models. Our key finding is that perturbations injected during the mixing step interval achieve comparable attack effectiveness without processing all timesteps. We develop a selective RGB channel strategy that applies attention maps to the red channel while using GradCAM-guided perturbations on green and blue channels. This design preserves image structure while maximizing misclassification in target models. TAIGen maintains visual quality with PSNR above 30 dB across all tested datasets. On ImageNet with VGGNet as source, TAIGen achieves 70.6\% success against ResNet, 80.8\% against MNASNet, and 97.8\% against ShuffleNet. The method generates adversarial examples 10× faster than existing diffusion-based attacks. Our method achieves the lowest robust accuracy, indicating it is the most impactful attack as the defense mechanism is least successful in purifying the images generated by TAIGen.
\end{abstract}    
\section{Introduction}
\label{sec:intro}
\setlength{\belowcaptionskip}{-4pt}
\begin{figure}[!htb]
\begin{center}
   \includegraphics[width=0.8\linewidth]{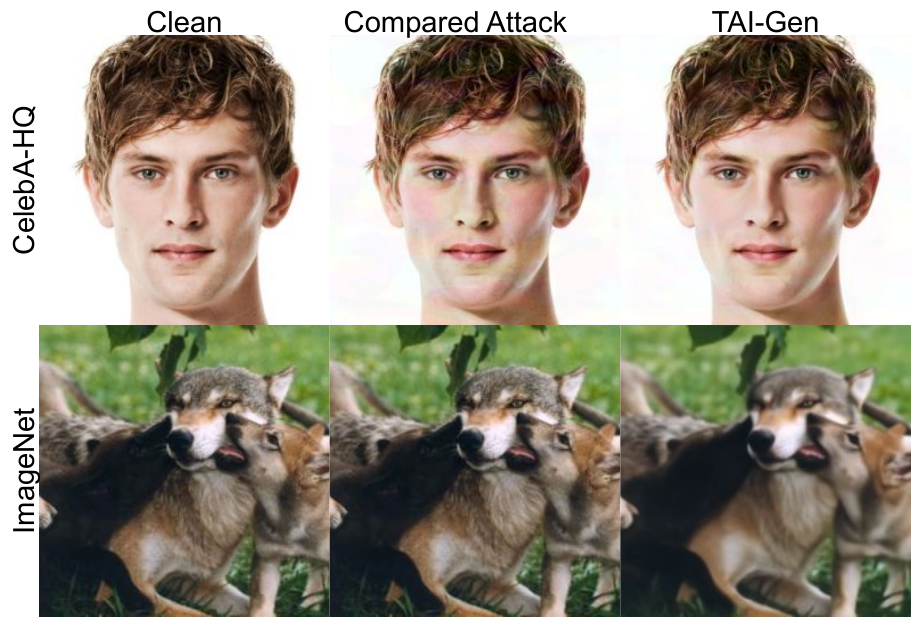}
\end{center}
   \caption{Images with their adversarially attacked counterparts. The first column is the clean images. The second is adversarial images generated using BPDA+EOT attack for the CelebA-HQ and AutoAttack for the ImageNet datasets. The third column is the adversarial images generated using our proposed TAIGen attack.}
\label{fig:abs}
\end{figure}
Diffusion models have quickly become a cornerstone in generative AI and their performance has surpassed the previous benchmarks in many tasks \cite{DBLP:conf/nips/DhariwalN21,DBLP:journals/pami/CroitoruHIS23,DBLP:journals/tkde/CaoTGXCHL24}. Diffusion models have shown to be successful in overcoming the issues faced by previous models such as training instabilities by GANs or the limited expressiveness by the autoregressive models. Additionally, the semantic space of diffusion models stays consistent over the timesteps\cite{kwon2023diffusion} which can be used for image manipulation without re-training. They have been utilized to generate adversarial examples, which can lead deep learning models to fail at their intended tasks. Some works have explored generating adversarial samples using diffusion models \cite{DBLP:conf/nips/KangSL23,DBLP:conf/nips/XueA0C23,DBLP:conf/aaai/LiuWP0H024}.  As can be seen in Figure \ref{fig:abs}, the different attack methods add unwanted artifacts to certain regions in faces(top row) and in multi-subject images(bottom row) affecting it's quality.
However, most of them are unrestricted in nature and utilize diffusion models in a white-box setting. They also utilize almost all the timesteps involved in the diffusion generation process which is inefficient. Finally, those methods have not fully utilized the underlying variation of the training dataset stored in the model which undermines the generative capability. \\
We propose a training-free method, TAIGen, that utilizes only a handful of timesteps to generate adversarial examples in a quick and robust manner by selectively modifying the RGB channels of the image.  We first find the significant timesteps by using key timestep \cite{zhu2023boundary} for convergence of Gaussian distribution to the data distribution. We further take advantage of attention maps as well as GradCAMs \cite{DBLP:conf/iccv/SelvarajuCDVPB17} to perturb the images. The RGB channels are modified selectively with this information as discussed later in detail. 
We showcase the performance of the proposed algorithm on three datasets – CIFAR-10, CelebA-HQ, and ImageNet datasets along with GradCAM analysis for the resulting adversarial images. 

\section{Related Work}
\begin{figure}[t]
\begin{center}
   \includegraphics[width=\linewidth]{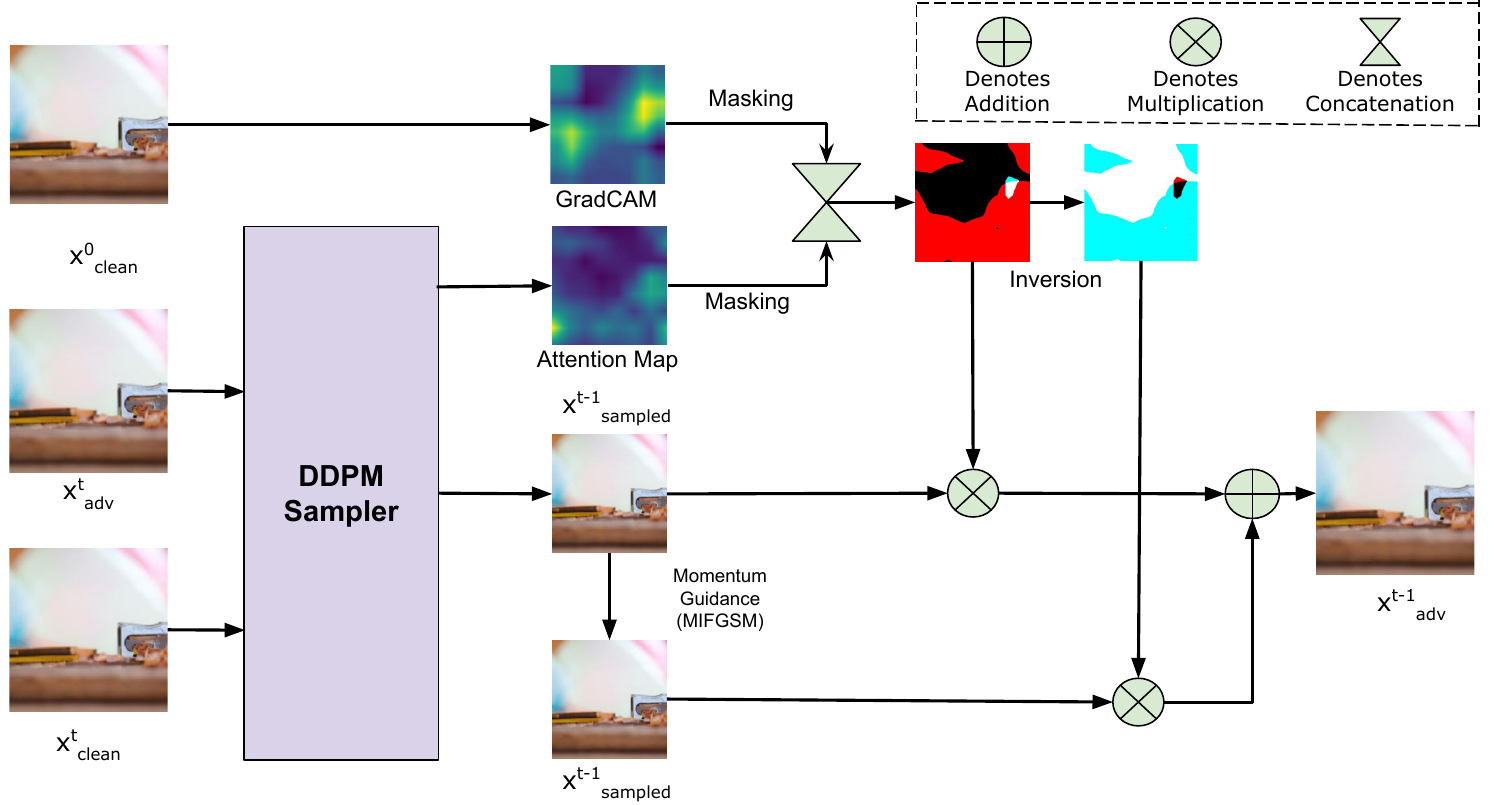}
\end{center}
   \caption{The proposed TAIGen adversarial attack where masked GradCAM and Attention Map are utilized for $N$ time steps in an iterative manner. We also use MI-FGSM \cite{DBLP:conf/cvpr/DongLPS0HL18} during the iteration to introduce momentum guidance. The DDPM Sampler denotes the noise prediction model and the backward process scheduler as a single unit.}
\label{fig:main}
\end{figure}

In this section, we review the current landscape for different adversarial attacks and purification methods.
\paragraph{Adversarial Attacks:} Adversarial Attacks continue to mislead deep learning models. Current research has shown the impressive success of adversarial attacks on a variety of models and modalities. The attacks can be categorized as white, gray, and black-box attacks based on the information available to the attacker about the source model. Further, they can be categorized based on their formulation as gradient-based, patch-based, and semantic-based methods. The adversarial perturbation can be solved as a constrained optimization problem, like FGSM \cite{DBLP:journals/corr/GoodfellowSS14}, PGD \cite{DBLP:conf/iclr/MadryMSTV18}, or CW \cite{DBLP:conf/sp/Carlini017} attacks, they are restricted by $l_p$-norm and utilize gradients of the source model. These attacks have been further improved by using parameters like momentum in MI-FGSM \cite{DBLP:conf/cvpr/DongLPS0HL18} or AdaMSI-FGM \cite{DBLP:conf/aaai/Long0LLZ24} attacks. Another way is to target semantic space to create adversarial examples for more natural samples like ColorFool \cite{DBLP:conf/cvpr/ShamsabadiSC20}. These adversarial samples can also be generated, as has been done using VAEs, GANs \cite{DBLP:conf/nips/SongSKE18, DBLP:journals/corr/abs-1904-07793, DBLP:conf/sp/KosFS18}, and more recently, diffusion models. With GANs, there are attacks like the Latent-HSJA attack \cite{DBLP:conf/eccv/NaJK22}, which generates unrestricted adversarial examples in a black-box setting, or the GMAA attack \cite{DBLP:conf/cvpr/LiHLZ0C23}, which creates a whole adversarial manifold over a point-wise generation. 

Owing to the poor quality of generated samples, especially on high-quality datasets, Diffusion models have been used instead of GANs to generate the adversarial samples. The models can utilize gradients as in AdvDiff \cite{10.1007/978-3-031-72952-2_6}, a training-free method, where PGD gradients are injected into the sampling of generative models giving two novel guidance techniques called adversarial and noise sampling guidance for adversarial example generation. AdvDiffuser \cite{10377423} also generates unrestricted adversarial examples and utilizes GradCAM \cite{DBLP:conf/iccv/SelvarajuCDVPB17} to retain important regions, only perturbing the less important ones. Even for face recognition systems, diffusion models have been used for generation. Adv-Diffusion \cite{DBLP:conf/aaai/LiuWP0H024} is one such attack that protects face identity by masking them using a reference image via a latent diffusion model. Another approach is Diff-AM \cite{Sun2024DiffAMDA}, which is a diffused-based adversarial makeup transfer method used for face protection by first having a text-guided makeup removal module, followed by image-guided adversarial makeup transfer. An ensemble of face recognition models is used as the attack strategy within this method. DiffAttack \cite{DBLP:conf/nips/KangSL23} is an attack that breaks diffusion-based adversarial purification methods by proposing a novel deviated reconstruction loss. These diffusion-based attacks are able to generate high-quality adversarial samples, however, with our work, we propose a training-free, less-time-consuming diffusion-based generation attack. \paragraph{Adversarial Defense and Purification:}With attack, several defense strategies have also been proposed in the literature, and similar to adversarial attacks, a variety of architectures can be utilized to defend against adversarial samples. Adversarial training still remains one of the most effective defense techniques. A self-supervised adversarial training method, CAFD \cite{DBLP:conf/iccv/Zhou0P0WYL21}, was proposed to remove adversarial noise in the class activation space. Another adversarial training method \cite{DBLP:conf/iccv/PoursaeedJYBL21} was proposed using generative models with a disentangled latent space to generate low, mid, and high-level changes in the samples. Several works have also studied adversarial robustness and detection in face recognition, including efforts on universal perturbation detection \cite{8698548}, attack-agnostic perturbation detectors \cite{9377649}, and robustness analysis of deep models against adversarial threats \cite{Goswami_Ratha_Agarwal_Singh_Vatsa_2018}. More recently diffusion models have been utilized for adversarial purification. DiffPure \cite{DBLP:conf/icml/NieGHXVA22} is one such method where a clean image is generated after diffusing the adversarial sample with a small amount of noise in the diffusion process. MimicDiffusion \cite{Song_2024_CVPR} approximates the generation process for a clean image and compares it with that of the adversarial sample to purify the image given two guidance techniques.

\section{TAIGen - The Proposed Attack}
Our work, Figure \ref{fig:main}, is based primarily on the Denoising Diffusion Probabilistic Model (DDPM); we therefore expand on it first in this section, followed by the problem formulation, the choice of a few sampling steps used in our algorithm, the selective information passed to different channels of the intermediate image, and the final methodology.  

\subsection{Background}
In this section, we introduce the Denoising Diffusion Probabilistic Model (DDPM) \cite{10.5555/3495724.3496298} which works via the inversion and sampling processes. Let us denote both the processes with the timestep index $t$ with $T$ as the diffusion length and each sample as $\{x_{t}\}_{t=0}^{t=T}$. DDPM constructs a discrete Markov chain $\{x_{t}\}_{t=0}^{t=T}$ with discrete time variables t following $p(x_{t}|x_{t-1}) = \mathcal{N}(x_{t};\sqrt{1-\beta_{t}}x_{t-1}, \beta_{t}\mathbf{I})$, where $\beta_{t}$ is the sequence of positive noise scales(e.g. linear, cosine scheduling) during inversion. Considering $\alpha_{t} = 1 - \beta_{t}$, $\Bar{\alpha_{t}} = \prod_{s=1}^{t}\alpha_{s}$, the sampling process in which the model learns to remove the Gaussian noise and predict \textbf{$x_{t-1}$} from a given latent variable \textbf{$x_{t}$} can be formulated as Equation \ref{eq:ddpm-1}.
\begin{equation}
    x_{t-1} = \frac{1}{\sqrt{\alpha_{t}}}(x_{t} - \frac{1-\alpha_{t}}{\sqrt{1-\Bar{\alpha_{t}}}}\epsilon_{\theta}(x_{t}, t)) + \sigma_{t}\textbf{z}
    \label{eq:ddpm-1}
\end{equation}

Here $\textbf{z} \sim \mathcal{N}(0,\mathbf{I})$, $\sigma_{t} = \eta \sqrt{\beta_{t}(1 - \Bar{\alpha}_{t-1})/(1-\Bar{\alpha}_{t})}$ is the standard deviation. $\eta$ is the level of stochasticity which is set to $1$ here. $\epsilon_{\theta}$ denotes the diffusion model parameterized by $\theta$ which is trained by minimizing the variational lower bound $\mathbb{E}_{q{(x_{0:T})}}$log($q(x_{1:T} | x_{0})$ / $p_{\theta}(x_{0:T})$) via the density gradient loss $\mathcal{L}_{d}$ as shown in Equation \ref{eq:ddpm-2}.
\begin{equation}
    \resizebox{0.9\hsize}{!}{$\mathcal{L}_{d} = \mathbb{E}_{t,\epsilon}\left[\frac{\beta_{t}^{2}}{2\sigma_{t}^{2}\alpha_{t}(1 - \Bar{\alpha_{t}})}||\epsilon - \epsilon_{\theta}(\sqrt{\Bar{\alpha}_{t}}x_{0} + \sqrt{1 - \Bar{ \alpha}_{t}}\epsilon, t) ||_{2}^{2}\right]$}
    \label{eq:ddpm-2}
\end{equation}



\subsection{Timestep Selection for Semantic Control}
\begin{figure*}[t]
\begin{center}
   \includegraphics[height=5cm, width=\textwidth]{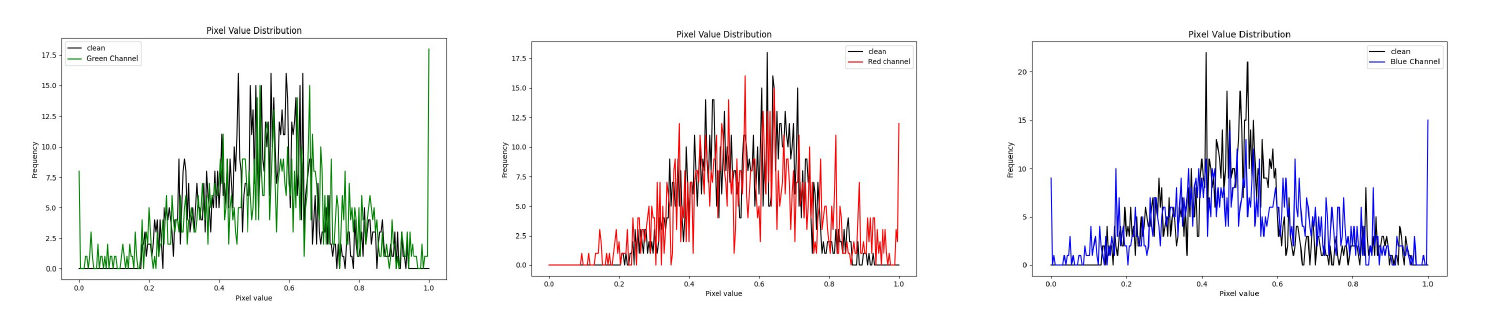}
   \caption{ Distribution of pixel values when attention maps are used in (a) Left: \textit{Green} channel, (b) Middle: \textit{Red} channel, (c) Right: \textit{Blue} channel. The black line indicates the distribution for the clean image, while the colored one is for the attacked image. Compared to others, variation from clean to attacked pixels is the least in the red channel.}
   \label{fig:fig2}
\end{center}
\end{figure*}

With TAIGen, we want to create a robust attack that will be visually similar to the original image whilst being computationally less expensive to create. Nonetheless, because the diffusion model evaluations are inherently deterministic, there is a fixed mapping from these noises to the final samples. Recent studies such as \cite{Ahn2024ANI} has shown that some noises are better than others in sampling. We thus choose a small interval($t << T$) of time steps during the sampling process instead of the entire $T$ steps. Additionally, this also increases the robustness when compared to a single timestep attack. 
\begin{figure}[t]
\begin{center}
   \includegraphics[width=0.8\linewidth]{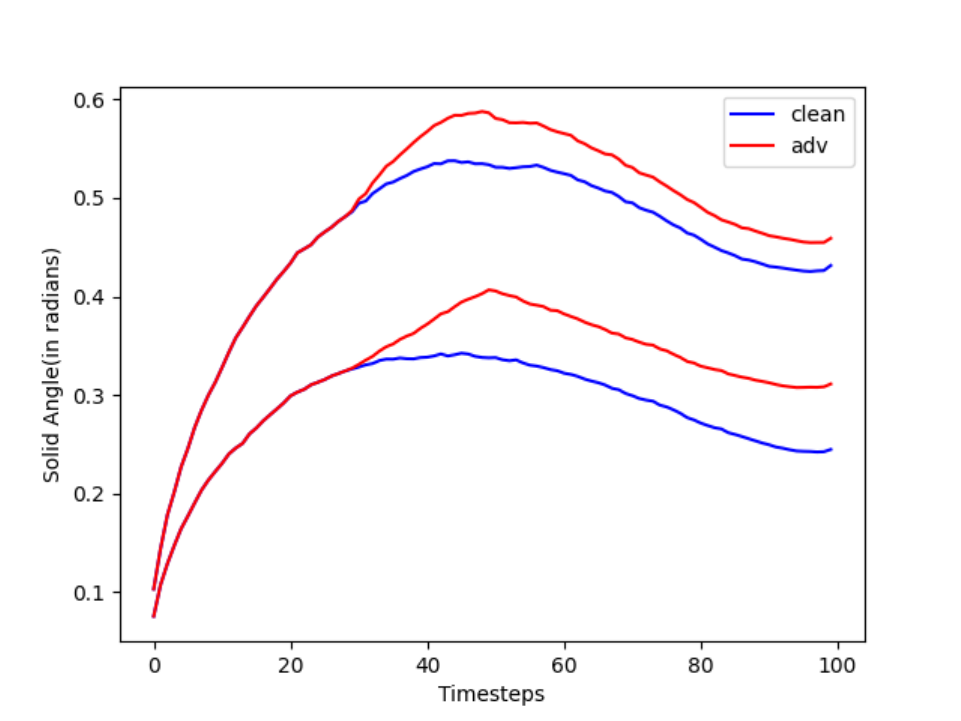}
\end{center}
\vspace{-5pt}
   \caption{Two samples from the CIFAR-10 dataset have been used. The Y-axis represents the solid angle between the latent variable $x_t$ and $\hat{x}_t$ during the forward and backward process respectively measured using the arcos function. The blue and red lines represent the convergence of original and adversarial reconstructions, respectively, highlighting the asymmetry and mixing step approximation.}
\label{fig:fig1}
\end{figure}
\begin{algorithm}[!t]
\caption{Adversarial Image Sampling}
\begin{algorithmic}[1]
\For {$t \in [T, T-1, \ldots, 1]$}
    \State $\epsilon_t, w_{t-1} \leftarrow \epsilon_\theta (\hat{x}_t, t)$
    \State $\hat{x}_{t-1} = \frac{1}{\sqrt{\bar{\alpha}_t}} \left( x_t - \frac{\beta_t}{\sqrt{1-\bar{\alpha}_t}} \epsilon_t \right) + \sigma_t z$
    \If {$(t \leq t_{start}) \quad \& \quad (t \geq t_{end})$}
        \State $\epsilon_t \leftarrow \epsilon_\theta (\hat{x}_{t}^{adv}, t)$
        \State $z_{0} \leftarrow \frac{1}{\sqrt{\bar{\alpha}_{t}}} \left( x_t - \frac{\beta_{t}}{\sqrt{1-\bar{\alpha}_{t}}} \epsilon_t \right) + \sigma_{t} z$
        \State $G_{t-1} \leftarrow \mathrm{1}(M > \Omega)$
        \State $W_{t-1} \leftarrow \mathrm{1}(w_{t-1} > \Phi)$
        \State $C_{t-1} \leftarrow W_{t-1}\oplus G_{t-1} \oplus G_{t-1}$
        \For {$i \in [0, 1, 2, ... I-1]$}
            \State $g_{i+1} = \mu g_i + \frac{\nabla_z J(z_i, y)}{\|\nabla_z J(z_i, y)\|_1}$
            \State $z_{i+1} = z_i + \alpha \cdot \text{Sign}(g_{i+1})$
        \EndFor
        \State \Return $z_I$
    \EndIf
    \State $\hat{x}_{t-1}^{adv} \leftarrow C_{t-1} \odot \hat{x}_{t-1}^{adv} + (1-C_{t-1}) \odot z_{I}$
    \If {$t == t_{end}$}
        \State $\hat{x}_{t} = \hat{x}_{t}^{adv}$
    \EndIf
\EndFor 
\State \Return $\hat{x}_{0}$
\end{algorithmic}
\label{alg}
\end{algorithm}
\paragraph{Small Interval over single-step: } As pointed out by \cite{zhu2023boundary}, a mixing step in the reverse process is defined as a key step($t_{mixing}$) in which the Gaussian distribution of $x_T$ converges to the final distribution of the input data in the denoising process under some pre-defined distance measures. Instead of resorting to manual checking of the quality of the final denoised images, we compute the radius of the latent encodings following the generative trajectory. Specifically, given a d-dimensional Gaussian centered at the origin with variance $\sigma^2$, for a point $x = (x_1,x_2,\dots ,x_d)$ chosen at random from Gaussian, the radius is the square root of the expected squared length of x:
\begin{equation}
    \sqrt{E(x_1^2 + x_2^2 + \dots + x_d^2)} = \sqrt{dE(x_1^2)} = \sqrt{d}\sigma
\end{equation}
The earliest step with the largest radius shift $\Delta r \approx 4$ is an approximation of the mixing step. By manipulating the latent space in this step, we can gain significant control over the distribution of the reconstructed sample. We utilize this mixing step to reduce the number of required sampling steps during adversarial image generation. Instead of using all the sampling steps during the reverse process, we only perform our proposed algorithm on the mixing step. However, empirically, we find that instead of focusing on a single step, a small interval of steps generates a more robust adversarial image. The generated image has the same image quality as compared to a single-step attack. Therefore, we choose the interval($N$) to be a hyperparameter in our experiments with $N\ll T$ with $T$ being the total number of timesteps in the sampling process. 
\paragraph{From all steps to small interval: } Figure \ref{fig:fig1} can help understand the reduction of sampling steps to a small interval. The point where the red line diverges from the blue line till the maxima is the size $N$. Contrary to previous beliefs \cite{kwon2023diffusion}, the asymmetric nature of the stochastic denoising and deterministic denoising process has a big advantage, i.e., control of the latent variable \textbf{$x_{t}$} in this interval via the $\epsilon$-space of the model can significantly control the semantics of the reconstructed image. We can also think of it as a perturbing latent variable near the region where it differs the most (hence maxima) from the actual inverted latent variable at that timestep. Therefore, we utilize this property to reduce the time needed to generate a sample, unlike methods that use almost all $T$ timesteps in the worst-case scenario \cite{10377423, DBLP:conf/aaai/LiuWP0H024}. Note that we find $t_{start}$ and $t_{end}$ empirically and these could change with the change in $T$. In any scenario however, $t_{start}$ can be found by calculating the radius of the latent variable. We provide a more detailed formulation of Figure \ref{fig:fig1} in the supplementary material section.


\subsection{Attention Maps: Leveraging h-space of DDPM Sampler}
The role of h-space has been shown to be critical in determining the semantics of the resultant image \cite{kwon2023diffusion} along with proving that it is solely responsible for controlling the semantic space \cite{Jeong2023TrainingfreeCI} of the images along with the skip connections. To effectively use this high-dimensional information, we extract the attention map: $W \in \mathcal{R}^{B\times HW \times HW}$, $HW$ being the number of tokens, from the self-attention module in the h-space or the middle-block of the U-Net architecture, which produces $\epsilon_{\theta}(x_{t},t)$ as it's output. We observe that if we consider all the attention maps from the encoder and decoder, our visual quality remains almost the same whilst the robust accuracy increases. Moreover, \cite{ddf36f4948414ac7921c0ff56cf538a7} showed that self-attention guidance can improve the stability and enhance the quality of the reconstructed image. Hence, we choose only to use the attention map from the middle block as it suffices for the majority of the latent space information.      
\begin{table*}[]
\resizebox{\linewidth}{!}{%
\begin{tabular}{|l|c|cccccccc|}
\hline
Model & Attack  &  Resnet  & EfficientNet & GoogleLeNet & MNASNet & MobileNet & SuffleNet & SqueezeNet & VGG \\
\hline
\multirow{4}{*}{ResNet} & PGD & \textbf{100.0} & 36.6 & 40.4 & 43.4 & 40.2 & 50.6 & 52.6 & 46.6 \\
       & AutoAttack & \textbf{100.0} & 34.6 & 40.2 & 42.8 & 40.8 & 50.0 & 54.0 & 48.4 \\
       & AdaMSI-FGM & \textbf{100.0} & 51.2 & 55.4 & 54.6 & 46.8 & 55.2 & 65.0 & 62.4 \\
       & \textbf{TAIGen (Ours)} & 68.8 & \textbf{62.0} & \textbf{60.6} & \textbf{87.4} & \textbf{61.2} & \textbf{94.2} & \textbf{96.41} & \textbf{91.6} \\
\hline
\multirow{4}{*}{GoogleLeNet} & PGD & 38.8& 33.8& \textbf{100.0}& 39.6& 41.2& 48.6& 49.8& 43.2 \\
       & AutoAttack & 38.8& 33.6& \textbf{100.0}& 41.6& 41.2& 49.2& 53.8& 44.8 \\
       & AdaMSI-FGM & 50.0 &42.8 & \textbf{100.0} & 49.0& 45.4 &53.6& 60.2 &54.2 \\
       & \textbf{TAIGen (Ours)} & \textbf{58.4} & \textbf{49.6} & 50.0 & \textbf{70.21} & \textbf{51.2} & \textbf{90.4} & \textbf{91.8} & \textbf{82.8} \\
\hline
\multirow{4}{*}{MobileNet} & PGD & 35.6& 36.6 &35.6& 53.4& \textbf{100.0}& 50.4& 51.8& 42.0 \\
       & AutoAttack & 33.6 &32.6 &35.0& 51.6 &\textbf{100.0}& 50.8 &49.8 &40.8 \\
       & AdaMSI-FGM & 43.0 &48.0 &43.4 &67.0 &\textbf{100.0}& 56.4& 59.4& 51.0 \\
       & \textbf{TAIGen (Ours)} & \textbf{64.2} & \textbf{62.4} & \textbf{57.0} & \textbf{81.8} & 73.4 & \textbf{92.99} & \textbf{94.6} & \textbf{89.0} \\
\hline
\multirow{4}{*}{VGG} & PGD & 43.8& 38.4 &39.2& 46.8& 43.0 &49.4 &59.0& \textbf{100.0} \\
       & AutoAttack & 42.4& 37.0 &39.8& 46.0 &41.6& 48.6& 58.6& \textbf{100.0} \\
       & AdaMSI-FGM & 58.2 &53.2& 52.2 &58.6 &48.6 &55.2& 70.0 &\textbf{100.0} \\
       & \textbf{TAIGen (Ours)} & \textbf{70.6} & \textbf{66.4} & \textbf{61.6} & \textbf{80.8} & \textbf{65.8} & \textbf{95.0} & \textbf{97.81} & 94.8 \\
\hline
\end{tabular}
}
\caption{ASR (\%) of adversarial attacks against eight models on the ImageNet dataset. The row depicts the source models, while the column depicts the target models.} 
\label{tab:tab2}
\end{table*}

\begin{table}[!t]
\begin{center}
\begin{tabular}{|l|ccc|}
\hline
Attribute  &  Method &  Clean Acc   & ASR \\
\hline
Eyeglasses  &  BPDA+EOT& 100 & 91.6  \\
            & \textbf{TAIGen (Ours)} & 100 & \textbf{100} \\
\hline
Smiling &  BPDA+EOT& 97.46 & 99.41  \\
            & \textbf{TAIGen (Ours)} & 97.46 & \textbf{100} \\
\hline
Gender &  BPDA+EOT& 100 & \textbf{100}  \\
            & \textbf{TAIGen (Ours)} & 100 & \textbf{100} \\
\hline
\end{tabular}
\caption{Comparison of BPDA+EOT attack on CelebA-HQ under $l_{\infty}$ bound and $\epsilon = 8/255$.} 
\label{tab:tab4}
\end{center}
\end{table}

\begin{table*}[]
\begin{center}
\begin{tabular}{|l|c|cccc|}
\hline
Attack Variant & Interval  &   ASR$(\uparrow)$  & FID$(\downarrow)$ & PSNR$(\uparrow)$ & SSIM$(\uparrow)$ \\
\hline
Clean & - & - & 0.09 & 32.20 & 88.66 \\
$l_{\infty}$($\epsilon = 4/255$) & 27-22         & \textbf{99.7}  & 0.62
&28.44 & 82.61 \\
$l_{\infty}$($\epsilon = 4/255$) & 21-17         & 99.6  & 0.55 & 29.19 & 83.93 \\
$l_{\infty}$($\epsilon = 4/255$) & 20-18 & 77.8  & 0.20
&30.83 & 87 \\
$l_{\infty}$($\epsilon = 8/255$) & 19   & 37.8  & \textbf{0.15} & \textbf{31.53} & \textbf{87.56} \\
\hline
\end{tabular}
\caption{Comparison of image quality and attack success rate with different time step intervals on the CelebA-HQ dataset with $T=100$. The results on the clean images is also reported for understanding the efficacy of our approach.} 
\label{tab:tab1}
\end{center}

\begin{center}
\begin{tabular}{|l|cc|c|cc|}
\hline
\multirow{2}{*}{Model}  &   \multirow{2}{*}{Classifier Acc.}  & \multirow{2}{*}{Clean Acc.} & \multirow{2}{*}{Epsilon} & ASR & ASR\\ \cline{5-6}
& & & &(w/o early stopping) & (w early stopping) \\
\hline
\multirow{2}{*}{WideResNet-28-10} & \multirow{2}{*}{95.53} &  \multirow{2}{*}{84.76} & 4/255 & 35.28 & \textbf{65.44} \\ 
                &  & & 8/255 & 63.55 & \textbf{97.47}\\
\hline
\multirow{2}{*}{WideResNet-70-16} & \multirow{2}{*}{95.53} &  \multirow{2}{*}{84.76} & 4/255 &31.09 & \textbf{57.81} \\
                &  & & 8/255 & 61.48 & \textbf{88.21} \\    
\hline
\end{tabular}
\caption{Comparison of different attack strengths against variations of WideResNet classifiers on the CIFAR-10 dataset. All the experiments are in the white-box setting for both with and without early stopping.} 
\label{tab:tab6}
\end{center}
\end{table*}


\subsection{Adversarial Classifier Guidance} 
The neurons in the convolutional layers look for high-level semantics and detailed spatial information. Grad-CAM uses the gradient information flowing into the last convolutional layer of the CNN to assign importance values to each neuron for a particular decision of interest. In order to obtain the class-discriminative localization map Grad-CAM, $M \in \mathbb{R}^{h\times w}$, we first compute the gradient of the score for class c, $y^c$ (before the softmax), with respect to feature map activations $A^k$ of a convolutional layer, i.e. $\frac{\delta y^c}{\delta A^k}$. Using this, we calculate the neuron importance weight $\alpha_k^c$ via:
\begin{equation}
    \alpha_k^c = \underbrace{\frac{1}{Z} \sum_i \sum_j} \underbrace{\frac{\partial y^c}{\partial A_{ij}^k}}
\end{equation}
Furthermore, we perform a weighted combination of forward activation maps, and follow it by a ReLU to obtain:
\begin{equation}
    M = \text{ReLU} \left( \sum_k \alpha_k^c A^k \right)
\end{equation}
Inspired by \cite{10377423}, we use the gradient weighted class activation mapping (GradCAM) from the last convolution layer of each source classifier $f$ corresponding to an input sample $x_{0}$. However, instead of producing the GradCAM($M$) corresponding to the true label $y$ relating $x_{0}$, we randomly choose another label $\hat{y}$ and localize the specific regions relating to that class in the feature space of the GradCAM. Another method of doing the same could be to negate the importance weight $\alpha_k^c$. However, this could lead to confusion for the classifier about which regions are contextually important to a class and which are just negative space. This lead us to choose a random label $\hat{y}$. We also want to add adversarial noise to the latent image due to which the perspective of the momentum-boosted iterative method \cite{DBLP:conf/cvpr/DongLPS0HL18}, which helps to escape local minima and stabilize the update direction, was introduced. This is done by approximating an adversarial example $x_{I}$, starting from a reference image $x_{0}$ as formulated in Equation \ref{eq:gradcam}

\begin{equation}
\begin{split}
    & g_{t+1} = \mu \cdot g_{t} + \frac{(\nabla_{x}J(f(x_{t}^{adv}), y)}{||\nabla_{x}J(f(x_{t}^{adv}), y)||_{1}} \\
    & x^{adv}_{t+1} = x^{adv}_{t} + \alpha \cdot sign(g_{t+1})
\end{split}
\label{eq:gradcam}
\end{equation}

Here, $J$ denotes the cross-entropy loss used for C-class classification problems, $\mu$ is the momentum factor and $\left\|\cdot\right\|_1$ is the $l_1$ norm. In our experiments, we make our adversarial examples satisfy the  $l_{\infty}$ bound by setting $\alpha = \epsilon / I$ with I being the number of iterations. Momentum, as observed empirically in our experiments, helped escape any local minima and reach the saddle point solution for minimizing the negative log-likelihood $\log p_{\theta}({\hat{x_{t}}})$ whilst maximizing the adversarial loss $J(f({x_{t}^{adv}}),y)$. However, as mentioned by \cite{Sun2019HeavyballAA} it might escape saddle points too. We therefore use a combination of masked GradCAM ($G_t$) and masked attention maps ($W_{t}$) to counter it using Equation \ref{eq:channel}
\begin{equation}
    C_{t} = W_{t}\oplus G_{t}\oplus G_t
    \label{eq:channel}
\end{equation}
\noindent where $\oplus$ denotes the concatenation operation along the RGB channel. By doing so, the range of values of $(z_I-\hat{x}_{t-1}^{adv})$ are scaled by a factor of $(1-C_{t-1})$. This provides two advantages: (1) $(1-W_{t})$ helps increase adversarial loss $J(f({x_{t}^{adv}}),y)$ due to it's inherent masking of relevant pixels and also pushes the sample away from clean sample convergence route during sampling and (2)$(1-G_t)$ provides the correct gradients of the classifer relating to that sample for minimizing $\log p_{\theta}({\hat{x_{t}}})$. Moreover, to make the $W_t \in \mathbb{R}^{B\times HW \times HW}$ match in dimensions with the GradCAM($G_{t} \in \mathbb{R}^{B\times h \times w}$), which is the same as the input sample, we used linear interpolation on $W_t$ to match it's dimensions with $G_t$. As discussed in earlier sections, TAIGen utilizes attention maps and GradCAM during the generation of adversarial images. This utilization is done carefully by selectively modifying the RGB channels. Modifying all channels with GradCAM reduced the image quality and modifying all of them with the attention map loses the randomization introduced using the GradCAM. As observed from Figure \ref{fig:fig2}, the combination of the two has two benefits. We can see that the variation of the pixel values in the red channel is the least from the target pixel values. Thus, we use $W_{t}$ on the Red channel as it preserves the quality. The saturation rate is better in this case, which implies that the $l_{\infty}$ norm is well-satisfied, leading to higher subtlety and transferability across different black-box models. Hence, the artifacts are removed and the semantic organization is maintained in $\hat{x_{t}}$.
\vspace{-7pt}
\subsection{Combing the Components: The Proposed Algorithm}
As discussed in earlier sections, TAIGen utilizes attention maps and GradCAM during the generation of adversarial images over the chosen $N$ interval of time steps. Hence, we have an iterative TAIGen attack as seen in Algorithm \ref{alg}. We use the GradCAM $M$ of a random non-objective class $\hat{y}$ to put more weight to the attacked regions which are present in $z_{i}$ 
and the attention maps to help the model remember the actual regions of interest for proper reconstruction. Thus, we use the $G_t$ twice and $W_{t}$ once. Figure \ref{fig:main} shows a visual demonstration of our algorithm and Algorithm \ref{alg} gives an overview of it.

\section{Experiments and Results}
We conduct experiments across multiple datasets and models to show the performance of our method. The experimental settings, results, and observations are detailed as follows.

\subsection{Experimental Setting}
\subsubsection{Datasets}To evaluate the proposed TAIGen algorithm, we evaluate it on the CelebA-HQ \cite{DBLP:journals/corr/abs-1710-10196}, CIFAR-10 \cite{krizhevsky2009learning} and ImageNet \cite{5206848} datasets. CIFAR-10 is a widely known image dataset that contains $60,000$ images of resolution $32 \times 32$. We consider $512$ images from the validation set ($10,000$ images) over $3$ different seeds following the protocol maintained by \cite{DBLP:conf/nips/KangSL23}. CelebA-HQ is a high-quality image dataset that contains $30,000$ images of resolution $1024 \times 1024$ which was made based on the CelebA dataset. We randomly select 512 images from this dataset. ImageNet is another widely known dataset that contains images of resolution $224 \times 224$. Following the protocol maintained by recent attacks \cite{DBLP:conf/aaai/Long0LLZ24}, we only consider randomly chosen 512 images or 1000 images from the validation set of ILRVSC 2012 \cite{ILSVRC15} for experiments.
\vspace{-2pt}
\subsubsection{Models} We use different models for our experiments. For experiments with ImageNet, we use ResNet-34 \cite{7780459}, EfficientNet-b0 \cite{Tan2019EfficientNetRM}, GoogleLeNet \cite{Szegedy2014GoingDW}, MobileNet-small \cite{Howard2019SearchingFM}, and VGG-11 \cite{Simonyan2014VeryDC} as source classifiers. While MNASNet-0-5 \cite{Tan2018MnasNetPN}, ShuffleNet-v2-x0-5 \cite{10.1007/978-3-030-01264-9_8}, and SqueezeNet-1-1 \cite{SqueezeNet} are the target models. For CIFAR-10. we use the variants of the WideResNet \cite{Zagoruyko2016WRN} model.

\subsubsection{Hyperparameters and Implementation Details} For CIFAR-10 and ImageNet, we consider $T=100$ and $N=20$. For CelebA-HQ, we consider $T=50$ and $N=3$. We use $\epsilon$ values as $4/255$, $8/255$, and $16/255$ to compare between different attack strengths. Additional details are available in the supplementary material. All the experiments were carried out on a single $32$ GB NVIDIA-V100 GPU. 

\subsubsection{Evaluation Metrics} In the case of CIFAR-10, we use the Robust Accuracy metric against the Diffpure \cite{DBLP:conf/icml/NieGHXVA22} purification technique. For CelebA-HQ and ImageNet, we use the Attack Success Rate (ASR) metric and report the top-1 accuracy. To evaluate the image quality of our generated samples, we also use the Peak-Signal-to-Noise Ratio (PSNR), Structural Similarity (SSIM), and Frechet Inception Distance (FID) metrics.

\subsection{Results and Analysis}
 \begin{table}[!t]
\begin{center}
\begin{tabular}{|l|c|}
\hline
Method            & Robust Acc. \\
\hline
SPSA*         & 81.29  \\
Square Attack*         & 81.68  \\
Joint Attack (Full)*        & 76.26  \\
Diff-BPDA*       & 75.00  \\  
AutoAttack*      & 70.64 \\
\hline
\textbf{TAIGen (Ours)} & \textbf{65.10}  \\
\hline
\end{tabular}
\caption{DDPM-based purification on CIFAR-10 dataset when compared with other attack methods on WideResNet-28-10 under $l_{\infty}$ norm and $\epsilon = 8 / 255$.} 
\label{tab:tab3}
\end{center}
\end{table}
\paragraph{Transferability with ImageNet:} We evaluate results for TAIGen along with other attack algorithms including PGD \cite{DBLP:conf/iclr/MadryMSTV18}, AutoAttack \cite{10.5555/3524938.3525144}, and AdaMSI-FGM \cite{DBLP:conf/aaai/Long0LLZ24}. We report the ASR in Table \ref{tab:tab2}. We observed that our method has a higher ASR rate than the other considered methods in almost all the cases for the black-box setting. While for the white-box setting, we do not perform as well. We conjecture that this is due to the fact that our algorithm balances all the aspects equally. Therefore, although we are using the gradients of the source classifier in the interval $N$, they are being integrated in a non-trivial manner in $x_{t}$ such that it can successfully confuse the other target classifiers into misclassification while in case of same target classifer as the source, some semantics are preserved in the GradCAM which guides the classifer. We also note that with the increase in the parameter space of the model, like in VGG-11, TAIGen performs very well. This is because small variations in the parameters of these models can significantly affect the prediction.
\paragraph{Imperceptibility with CelebA-HQ: }
To evaluate our attack on face recognition systems, we use the CelebA-HQ dataset and BPDA+EOT attack \cite{DBLP:conf/iclr/HillMZ21} to compare our algorithm against as reported in Table \ref{tab:tab4}. We use the attribute prediction task to evaluate our results with target attributes selected as \textit{Eyeglasses}, \textit{Smiling}, and \textit{Gender}. We note that TAIGen performs better than the considered attack method. Further, we evaluate the image quality of our attack over different time intervals on the attribute-wise robust target classifier proposed by \cite{chai2021ensembling}. We report ASR along with the image quality metrics in Table \ref{tab:tab1}. We use only the gender-attribute prediction task for this experiment. As shown by the ASR metric, $t_{mixing}=19$ is an effective mixing step and our comparison with a single-step version of our attack. It sufficiently brings down the accuracy while maintaining the semantics of the image to almost near perfection. As we increase the interval size around the mixing step, we see the accuracy reduce even further with almost 100\% success rate with just $N=5$ steps around the mixing step. The second row shows us that as long as we are around the mixing step, we will still have a high success rate. Note that the mixing step is approximated via the method explained earlier and may vary a bit for different batches of samples. For intrusion in biometric systems, this procedure can be extended to similar datasets.

\paragraph{Robustness to Purification techniques:} Purification methods have been used to evaluate the robustness of several attack methods. To compare the robustness of TAIGen, we evaluate it on the DiffPure \cite{DBLP:conf/icml/NieGHXVA22} purification method. We compare it against SPSA \cite{DBLP:conf/icml/UesatoOKO18}, Square attack \cite{10.1007/978-3-030-58592-1_29}, Diff-BPDA\cite{Blau2022ThreatMA}, Joint attack\cite{e294fc2db268437e86892bc9b612d302}, and AutoAttack \cite{10.5555/3524938.3525144} $l_{\infty}$ threat model. SPSA attack approximates the gradients by randomly sampling from a predefined distribution and using the finite-difference method. Square attack heuristically searches for adversarial examples in a low-dimensional space with the constraints of perturbation patterns. Joint attack leverages the classifier gradients and the difference between the input and the purified samples. The results are reported in Table \ref{tab:tab3}, where our method outperforms all the considered methods in terms of robust accuracy.  

\begin{figure*}[t]
\begin{center}
   \includegraphics[width=0.9\linewidth]{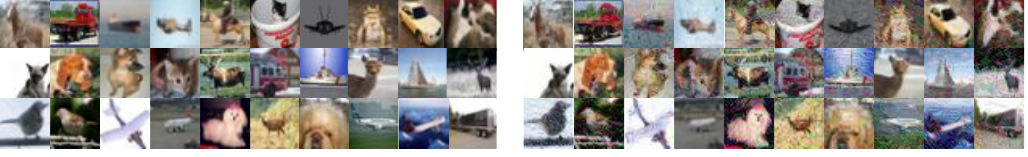}
\end{center}
\vspace{-7pt}
   \caption{A set of few original (left) and attacked images (right) of the CIFAR-10 dataset attacked using TAIGen.}
\label{fig:fig6}
\end{figure*}

\begin{figure}[!t]
\begin{center}
   \includegraphics[height=8cm, width=0.6\linewidth]{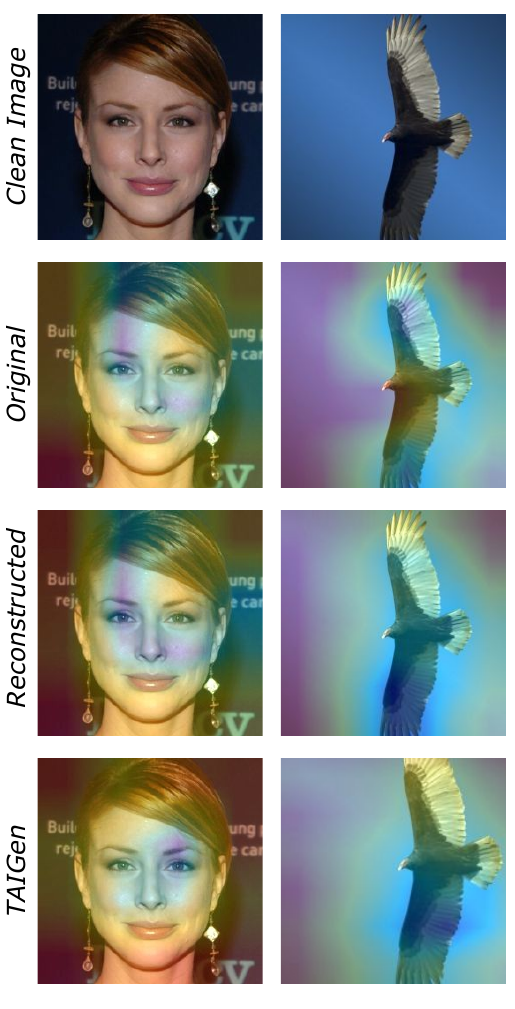}
\end{center}
\vspace{-7pt}
   \caption{Heatmaps produced by the classifiers on the CelebA-HQ (first column) and ImageNet (second column) datasets. The second row shows the GradCAM of the original image produced by the classifier, while the third and fourth rows show GradCAM of the DDPM reconstructed image and the TAIGen-based DDPM reconstructed image.}
\label{fig:attn-celebhq}
\end{figure}

\subsection{Ablation and Analysis}

\paragraph{Early Stopping:} 
\begin{table}[t]
\begin{center}
\begin{tabular}{|l|c|}
\hline
Attack  &  Time(in seconds)\\
\hline
ACA*  &  125.33 \\
\textbf{TAIGen (Ours)}(w/o early stopping) & 12 \\
\textbf{TAIGen (Ours)}(w early stopping) & \textbf{6.6} \\
\hline
\end{tabular}
\caption{Comparison of ACA attack which creates adversarial images using the ImageNet dataset with MobileNet-V2 as the source and target classifier. $*$ indicates that the values have been taken from their respective papers in the reported tables.} 
\label{tab:tab7}
\end{center}
\end{table}
Using early stopping with our method significantly reduces generation time, as shown in Table \ref{tab:tab7}, outperforming the ACA attack \cite{10.5555/3666122.3668375}. On the CIFAR-10 dataset, with WideResNet-28-10 and WideResNet-70-16 as robust classifiers, early stopping increases ASR by $\sim30\%$ 
(Table \ref{tab:tab6}). This demonstrates that early stopping effectively reduces high-frequency noise due to less cumulative scaling of the latent variable $x_{t}$	whilst speeding up generation. Further details can be found in the supplementary material.


\paragraph{GradCAM Analysis:} We pull the weights from the last convolution layer of the attribute classifier in the case of CelebA-HQ and Resnet-34 in the case of ImageNet to create the GradCAMs. We observe that TAIGen avoids focusing on regions sensitive to gender classification and localizes areas completely irrelevant to the classification, as seen in Figure \ref{fig:attn-celebhq}. A major reason for the shift in semantic localization can be attributed to the masking operation we are performing on the GradCAM and attention maps in our algorithm. This is because we are attending to each of the channels disproportionately which leads to improper scaling by the coefficient of $\epsilon_{\theta}(x_{t},t)$ across the timesteps. Another important point is that in the white-box setting, our method can easily displace points of interest when there are fewer important features. However, when the complexity of the feature of an image increases, as in the case of ImageNet, $W_{t}$ masks many regions, which prevents the addition of perturbations to them. 
\paragraph{Visualization:} Figures \ref{fig:fig6} and \ref{fig:attn-celebhq} show the visualizations of our attacked images. In the case of CIFAR-10, we keep $l_{\infty}=8/255$ to provide the worst-case scenario of our method in very low-resolution images. However, most biometric datasets such as face and iris are of high resolution and we notice that we obtain high-quality adversarial images in high-resolution images, thereby increasing applicability. Additionally, our method does not add any unnecessary features, such as a change in shape or texture, to the original image. This restricts the diffusion models to add unwanted features to the reconstructed images by maintaining the sanctity of the h-space through the attention maps. 

\section{Conclusion}
Our proposed TAIGen method is a black-box, few-step, iterative diffusion-based attack method that selectively modifies the channels in the target image. Through experiments on different datasets and models, we show the robust performance of our attack in terms of performance, imperceptibility, and transferability. We observed that our method performs well in a black-box setting and thus has better transferability. The method generates high-quality adversarial images, specifically in biometric datasets. Although our method has some limitations, such as low performance in a white-box setting, we are able to generate a robust adversarial image that performs better than many state-of-the-art methods and showcases the effectiveness of diffusion models. In future work, we hope to improve the attack to withstand stronger purification techniques and address the issue of deterioration of quality in low-resolution images.
\section{Acknowledgement}
The authors acknowledge the support of IndiaAI and Meta
through the Srijan: Centre of Excellence for Generative AI. 
{\small
\bibliographystyle{ieeenat_fullname}
\bibliography{main}
}
\maketitlesupplementary



\section{Experimental Setting}
\label{sec:Exp2}
In addition to the details presented in the main paper, the additional experimental settings are detailed below. 

\subsection{Hyperparameter Details:}
\label{sec:Details2}
The values of $\Omega$ and $\Phi$ have been set to $0.9$. Additionally, we used linear beta time scheduling ($\beta_{t}$) with the standard DDPM \cite{10.5555/3495724.3496298} configurations in all the cases. 

\paragraph{ImageNet:} We used the Guided Diffusion model \cite{DBLP:conf/nips/DhariwalN21} from OpenAI on this dataset. The batch size was set to 4. We used 512 images, each of size $224\times 224$. The values of $t_{start}$ and $t_{end}$ here are $80$ and $60$ respectively. We also kept $\epsilon = 4/255$ when comparing against PGD, AutoAttack, and AdaMSI-FGM. While testing against ACA, we used the MobileNet-V2 as the source and target model and 1000 images were chosen at random. The reported time is in seconds/image. The momentum factor was set to $1.2$, and I was fixed at $20$ iterations in both of these experiments. We perform a $10$-class classification and randomly choose a class from the top-$5$ classes next in line to the true label class for creating the Grad-CAM.

\paragraph{CelebA-HQ:} Whilst testing against BPDA+EOT, we kept $\epsilon = 8/255$, $t_{start}=20$ and $t_{end}=18$ and batch size as $4$. Since there are only two classes in this case, we choose the class that does not correspond to the true label in this case for creating the Grad-CAM.

\paragraph{CIFAR-10:} We set batch size as $8$, $I=50$, $512$ as the number of images, $t_{start}=80$ and $t_{end}=55$. We also fixed $\epsilon=4/255$ when comparing the robust accuracy against SPSA, Square Attack, Joint Attack (Full), Diff-BPDA and Auto-Attack. The momentum factor is $1.8$. We perform a 1000-class classification and choose a class randomly from the top-500 classes, which are next in line to the true label class for creating the Grad-CAM.

\subsection{Additional Experimental Results:}
\label{sec:exp3}

\begin{table}[]
\begin{center}
\begin{tabular}{|l|cc|}
\hline
  & PSNR & SSIM \\
\hline
Clean                            & 24.16 & 81.55 \\
\hline
w/o early stopping                &  22.42 & 75.39\\
w early stopping              &  \textbf{23.28}  & \textbf{77.12}\\
\hline
\end{tabular}
\caption{PSNR and SSIM values on the CIFAR-10 dataset. The model used is the standard WideResNet-28-10 with $\epsilon=4/255$.} 
\label{tab:table1}
\end{center}
\end{table}

\begin{algorithm}[!t]
\caption{Adversarial Image Generation with Early Stopping}
\begin{algorithmic}
\Function {TAIGen}{\small Input Image $x_0$, Noise Schedule $\beta_{1:T}$, Target Classifier $f$, Grad-CAM $g_{\text{CAM}}$, Diffusion U-Net Model $\epsilon_\theta$, Ground Truth $y$, Adversarial Iterations $I$, Momentum Factor $\mu$, Step Size $\alpha$, Cross-Entropy loss $J$}
\State {\textbf{Ensure:}} Adversarial Image $x_{\text{adv}}$
\State {\textbf{Initialize:}} $x_T \leftarrow x_0 \cdot \sqrt{\bar{\alpha}_T} + \sqrt{1-\bar{\alpha}_T} \cdot \epsilon$, $\epsilon \sim \mathcal{N}(0, I)$
\State $\hat{x}_{t}^{adv} \leftarrow \hat{x}_{t_{start}}$, $\sigma_t^{2} \leftarrow \beta_{t}$, $z \sim \mathcal{N}(0, I)$, $N = t_{start} - t_{end}$

\For {$t \in [T, T-1, \ldots, 1]$}
    \State $\epsilon_t, w_{t-1} \leftarrow \epsilon_\theta (\hat{x}_t, t)$
    \State $\hat{x}_{t-1} = \frac{1}{\sqrt{\bar{\alpha}_t}} \left( x_t - \frac{\beta_t}{\sqrt{1-\bar{\alpha}_t}} \epsilon_t \right) + \sigma_t z$
    \State $x_{0} \leftarrow \frac{1}{\sqrt{\bar{\alpha}_{t}}}(x_{t} - \sqrt{1 - \bar{\alpha}_{t}}\epsilon_{t})$
    \If {$\text{arg max} f(x_{0}) \neq y$}
    \State $\hat{x}_{0} = x_{0}$
    \State \textbf{break}
    \ElsIf {$(t \leq t_{start}) \quad \& \quad (t \geq t_{end})$}
        \State $\epsilon_t \leftarrow \epsilon_\theta (\hat{x}_{t}^{adv}, t)$
        \State $z_{0} \leftarrow \frac{1}{\sqrt{\bar{\alpha}_{t}}} \left( x_t - \frac{\beta_{t}}{\sqrt{1-\bar{\alpha}_{t}}} \epsilon_t \right) + \sigma_{t} z$
        \State $G_{t-1} \leftarrow \mathrm{1}(g_{CAM} > \Omega)$
        \State $W_{t-1} \leftarrow \mathrm{1}(w_{t-1} > \Phi)$
        \State $C_{t-1} \leftarrow (W_{t-1} \oplus G_{t-1}\oplus G_{t-1})$
        \For {$i \in [0, 1, 2, ... I-1]$}
            \State $g_{i+1} = \mu g_i + \frac{\nabla_z J(z_i, y)}{\|\nabla_z J(z_i, y)\|_1}$
            \State $z_{i+1} = z_i + \alpha \cdot \text{Sign}(g_{i+1})$
        \EndFor
        \State \Return $z_I$
    \EndIf
    \State $\hat{x}_{t-1}^{adv} \leftarrow C_{t-1} \odot \hat{x}_{t-1}^{adv} + (1-C_{t-1}) \odot z_{I}$
    \If {$t == t_{end}$}
        \State $\hat{x}_{t} = \hat{x}_{t}^{adv}$
    \EndIf
\EndFor 
\State \Return $\hat{x}_{0}$
\EndFunction
\end{algorithmic}
\label{alg1}
\end{algorithm}

\begin{table}[t]
\centering
\resizebox{\linewidth}{!}{%
\begin{tabular}{|l|c|c|}
\hline
\textbf{Model}                            & \textbf{Clean Accuracy (\%)} & \textbf{ASR (\%)} \\ \hline
R50 \cite{7780459}                     & 76.52                        & 96.10             \\ \hline
R50 \cite{DBLP:journals/corr/abs-2007-08489}                 & 64.02                        & 57.23             \\ \hline
WRN-50-2 \cite{DBLP:journals/corr/abs-2007-08489}           & 68.46                        & 54.11             \\ \hline
R50 \cite{robustness}              & 62.56                        & 57.43             \\ \hline
ViT-B \cite{moadversarial}           & 68.38                & 49.22     \\ \hline
\end{tabular}
}
\caption{Accuracy and ASR on the ImageNet dataset with $\epsilon = 4/255$ using the WideResNet-50-2 as the source model.}
\label{tab:table2}
\end{table}

To compare the image quality against the two variations (that is, with and without early stopping) of our algorithm, we randomly choose 512 images from the validation set of CIFAR-10 and report the PSNR and SSIM values in Table \ref{tab:table1}. We also showcase our algorithm with early stopping in Algorithm \ref{alg1}. We observe that image quality is retained more with early stopping. However, one limitation of using this is that the images become more susceptible to purification algorithms like Diffpure \cite{nie2022DiffPure}. This is because, with early stopping, the images are not as robust as those that have undergone more iterations of adding adversarial noise. Thus, they are easier to purify. Since we present a black-box attack, we use the WideResNet-50-2 classifier \cite{DBLP:journals/corr/abs-2007-08489} as the source model and test the efficacy on various adversarially trained target models, ResNet-50 \cite{robustness}, ResNet-50\cite{DBLP:journals/corr/abs-2007-08489} and WideResNet-50-2 \cite{DBLP:journals/corr/abs-2007-08489}, ViT-B \cite{moadversarial} and a standard ResNet-50 model \cite{7780459}. The values for this experiment are reported in Table \ref{tab:table2}.

\section{Finding Mixing Step}

\cite{zhu2023boundary} has explained in detail about the mixing step. We give a brief explanation of the same. We first define the radius of a high-dimensional Gaussian space. Mathematically, it can defined as $r = \sigma\sqrt{d}$. Now let us take a point in this vector space, $x = (x_{1},x_{2},\dots,x_{d})$, chosen at random from Gaussian, the square root of the expected square length of x is formulated in Equation \ref{eq:eq1}.

\begin{equation}
    \sqrt{\mathrm{E}(x_{1}^{2}+x_{2}^{2}+\dots+x_{d}^{2})} = \sqrt{d\mathrm{E}(x_{1}^{2})} = \sqrt{d}\sigma
    \label{eq:eq1}
\end{equation}

Equation \ref{eq:eq1} is used to find the radius of our sampled latent variables at each time step. We also define the total variation distance which will be used in the further proofs in Equation \ref{eq:eq2}.

\begin{equation}
    ||\mu - \tau||_{TV} = \frac{1}{2}\sum_{x \in \mathcal{X}} |\mu(x) - \tau(x)|
    \label{eq:eq2}
\end{equation}

Here $\mu$ and $\tau$ are two probability distributions on $\mathcal{X}$. Next, we define the quantity $\sigma_{t}(x,y)$ for an irreducible transition matrix $P$ with stationary distribution $\pi$ in Equation \ref{eq:eq3} and $d^{(p)}$ distance in Equation \ref{eq:eq4}.

\begin{equation}
    \sigma_{t}(x,y) = \frac{P^{t}(x,y)}{\pi(y)}
    \label{eq:eq3}
\end{equation}

\begin{equation}
    d^{p}(t) := \text{max}_{x \in \mathrm{X}}||\sigma_{t}(x,.) - 1||_{p}
    \label{eq:eq4}
\end{equation}

Replacing the above notations with the ones from a standard DDPM model, we get Equations \ref{eq:eq5} and \ref{eq:eq6}.

\begin{equation}
    d^{1}(t) := \text{max}_{x \in \mathrm{X}}||\sigma_{t}(x,.) - 1||_{1}  
    \label{eq:eq5}
\end{equation}

and,

\begin{equation}
    \sigma_{t}(x,y) = \frac{P^{t}(x,y)}{\pi(y)} = \frac{x \sim \mathcal{N}(x_{t};\sqrt{1-\beta_{t}}x_{t-1},\beta_{t}\mathrm{I})}{y \sim \mathcal{N}(0,\mathrm{I}_{d})}
    \label{eq:eq6}
\end{equation}

Mathematically, we can define the mixing time via Equation \ref{eq:eq7}

\begin{equation}
    t_{mix}^{(1)}(\epsilon) := \text{inf}\{t\geq 0; d^{1}(t) \leq \epsilon\}
    \label{eq:eq7}
\end{equation}

Taking $\epsilon=1/2$, we get Equation \ref{eq:eq8}

\begin{equation}
    t_{mix}^{(1)}(\epsilon) := \text{inf}\{t\geq 0; d^{1}(t) \leq \frac{1}2 \}    \label{eq:eq8}
\end{equation}

Replacing Equation \ref{eq:eq8} with Equation \ref{eq:eq6}, we get Equation \ref{eq:eq9}.

\begin{equation}
    \text{max}_{x \in \mathcal{X}}||\frac{x \sim \mathcal{N}(x_{t};\sqrt{1-\beta_{t}}x_{t-1},\beta_{t}\mathrm{I})}{y \sim \mathcal{N}(0,\mathrm{I}_{d})} - 1|| \leq \frac{1}{2}
    \label{eq:eq9}
\end{equation}

Using Equation \ref{eq:eq2}, we can substitute in Equation \ref{eq:eq9} which gives us the approximation in Equation \ref{eq:eq10}.

\begin{equation}
    ||x \sim \mathcal{N}(x_{t};\sqrt{1-\beta_{t}}x_{t-1},\beta_{t}\mathrm{I})|| \leq 4
    \label{eq:eq10}
\end{equation}

Equation \ref{eq:eq10} searches for the mixing step where the Gaussian radius changes by an amount of \textbf{4} units.

\section{Solid Angle using Quaternions}
\label{sec:solidangle}
To find the similarity between the latent variable $x_{t}$ during the forward process and $\hat{x}_{t}$ during the backward process, we use the cosine similarity between these two variables. Specifically, we consider the vector or quaternion $q_{1}$ and $q_{2}$ and the cosine similarity is found using Equation \ref{eq:eq11}.

\begin{equation}
   \Omega = \cos^{-1}\left(\frac{q_{1}}{||q_{1}||_{2}}\cdot \frac{q_{2}}{||q_{2}||_{2}}\right)
   \label{eq:eq11}
\end{equation}

As observed from the graph (Figure $3$ of the main paper), it forms a concave curve. Intuitively, we can say that while the Gaussian distribution doesn't converge with the stationary distribution, the solid angle or dissimilarity increases. At a certain point, in this case the maxima, they converge and then the similarity increases between the two vectors. Thus, via this analysis, it is evident that the inversion and the sampling processes are \textbf{not} symmetric. Now our intuition lies in utilizing this fact and perturbing the $\hat{x}_{t}$ around the time step corresponding to the maxima. If instead, we perturbed at any other time step, the dissimilarity would again increase till it reached a maximum (which would not be the mixing step) and then decrease. In the latter case, we empirically found the artifacts are visible in the reconstructed image.


\end{document}